%% file: main.tex
\documentclass[10pt,twocolumn,letterpaper]{article}

\usepackage[pagenumbers]{iccv} 


\definecolor{iccvblue}{rgb}{0.21,0.49,0.74}
\usepackage[pagebackref,breaklinks,colorlinks,allcolors=iccvblue]{hyperref}

\def\paperID{13146} 
\def\confName{ICCV}
\def\confYear{2025}

\title{CoStoDet-DDPM: Collaborative Training of Stochastic and Deterministic Models Improves Surgical Workflow Anticipation and Recognition}

\author{Kaixiang Yang$^{\dag}$, Xin Li$^{\dag}$,
Qiang Li, Zhiwei Wang$^{*}$\\
Wuhan National Laboratory for Optoelectronics, Huazhong University of Science and Technology\\
$^\dag$: Co-first authors, $^*$: Corresponding authors.\\
{\tt\small \{kxyang, lixin2023, liqiang8, zwwang\}@hust.edu.cn}
}

\begin{document}
\maketitle
\input{sec/0_abstract}    
\input{sec/1_intro}
\input{sec/2_methods}
\input{sec/3_experiment}
\input{sec/4_discussion}
{
    \small
    \bibliographystyle{ieeenat_fullname}
    \bibliography{main}
}

\input{sec/X_suppl}

\end{document}

%% file: sec/0_abstract.tex
\begin{abstract}
Anticipating and recognizing surgical workflows are critical for intelligent surgical assistance systems. However, existing methods rely on deterministic decision-making, struggling to generalize across the large anatomical and procedural variations inherent in real-world surgeries.
In this paper, we introduce an innovative framework that incorporates stochastic modeling through a denoising diffusion probabilistic model (DDPM) into conventional deterministic learning for surgical workflow analysis. At the heart of our approach is a collaborative co-training paradigm: the DDPM branch captures procedural uncertainties to enrich feature representations, while the task branch focuses on predicting surgical phases and instrument usage.
Theoretically, we demonstrate that this mutual refinement mechanism benefits both branches: the DDPM reduces prediction errors in uncertain scenarios, and the task branch directs the DDPM toward clinically meaningful representations. Notably, the DDPM branch is discarded during inference, enabling real-time predictions without sacrificing accuracy.
Experiments on the Cholec80 dataset show that for the anticipation task, our method achieves a 16\% reduction in eMAE compared to state-of-the-art approaches, and for phase recognition, it improves the Jaccard score by 1.0\%. Additionally, on the AutoLaparo dataset, our method achieves a 1.5\% improvement in the Jaccard score for phase recognition, while also exhibiting robust generalization to patient-specific variations. 
Our code and weight are available at \url{https://github.com/kk42yy/CoStoDet-DDPM}.

\end{abstract}

%% file: sec/1_intro.tex
\section{Introduction}
\label{sec:intro}

Artificial intelligence-assisted surgical workflow analysis is beneficial for the efficiency and safety of Robotic-Assisted Surgery (RAS)~\cite{MIA22_Anti,maier2022surgical}. Anticipation and recognition are two of the most critical tasks in this analysis.
The anticipation task refers to forecasting the occurrence of surgical instruments and phases before they manifest~\cite{MICCAI20rethinking,forestier2017automatic}, while the recognition task~\cite{SurPhaseDef} involves identifying the current surgical phase.
Accurate anticipation and recognition improve patient safety, reduce surgical errors, and foster communication and collaboration within the surgical team~\cite{ORCommunication,phaseeffect1,phaseeffect2,phaseoptimizeeffect,phasealert,phasedecmake}.

\begin{figure}
  \centering
   \includegraphics[width=\linewidth]{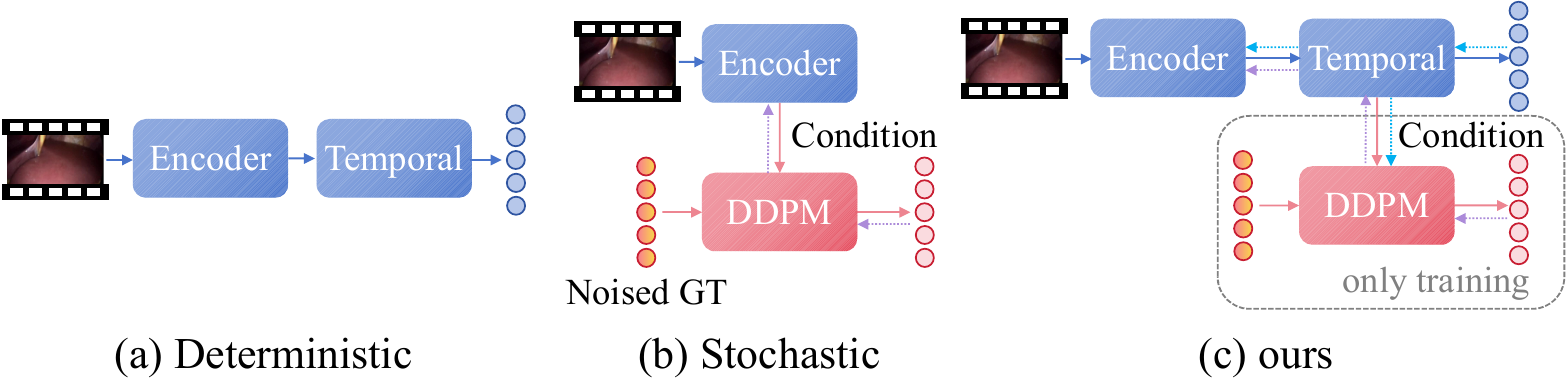}

   \caption{Different paradigms. 
   (a) The deterministic workflow used in previous methods fails to account for the individual patient variations. 
   (b) The conditional DDPM used in natural action segmentation is difficult to meet the real-time requirements and, when applied to clinical settings, lacks clinical consistency.
   (c) Our proposed co-training approach combines deterministic and stochastic models, allowing DDPM to be discarded during inference. 
   The dashed arrows indicate the gradient back-propagation.}
   \label{fig:paradigms}
\end{figure}

As shown in Figure~\ref{fig:paradigms}(a), current deep learning-based research on anticipation~\cite{forestier2017automatic,MICCAI20rethinking,rsdnet_anti_TMI18,catanet_anti_miccai21,MIA22_Anti,TimeLSTM_anti_miccai17,BHI_anti22,ADA_AntiSOTA,yuan_anti_miccai21} and recognition~\cite{ref11multi,ref12multi,ref13multimtrcnet,ref14lstmSV-RCNet,ref15lstm,ref16lstm,ref17lstmTMRNet,ref18lstmBNPitfalls,ref19TCNTECNO,ref20TransOpera,ref21TransTransSVNet,ref22Translovit,ref23TransCMTNet,ref24TransSkit,ref25DACAT,ref26Surgformer} typically utilizes a frame encoder, such as ResNet~\cite{ref10resnet} or Transformer~\cite{ref6Transformer,ref7ViT}, to extract embeddings for each individual frame. These embeddings are then processed by a temporal modeling component, such as Long Short-Term Memory (LSTM)~\cite{ref4LSTM} or Temporal Convolutional Networks (TCN)~\cite{ref5TCN}, to make final predictions based on the spatio-temporal features.
However, existing methods mostly rely on deterministic decision-making, establishing a fixed mapping from visual inputs to surgical phase or instrument predictions. These approach lead to an inherent limitation: \emph{collapse to dominant patterns}, making them vulnerable to anatomical anomalies or procedural deviations. As a result, they regress toward population-level averages rather than adapting to case-specific contexts. This is particularly detrimental in surgical workflow analysis, as evidenced by the unsmooth predictions and disordered intra-stage recognition observed in previous methods.


In recent years, the stochastic approach has garnered increasing attention in fields like image generation and robot behavior modeling, with notable models such as Denoising Diffusion Probabilistic Models (DDPM)~\cite{DDPM,DDIM}. These models tackle key challenge faced by deterministic methods through \emph{probabilistic sampling} and \emph{multi-modal outputs}, as shown in Figure~\ref{fig:paradigms}(b). These characteristics have made DDPM particularly successful in domains where variability and uncertainty are critical factors, such as generating diverse images or predicting robot actions.
Inspired by these successes, we are the first to explore the potential of stochastic methods in surgical workflow analysis, aiming to enhance the model's ability to handle patient-specific variations. However, directly applying DDPM to surgical workflow analysis presents significant challenges. First, the generated outputs may be anatomically implausible, violating clinical consistency and disrupting the logical flow of surgical procedures. Second, DDPM struggles to meet the real-time processing requirements of clinical settings, a crucial limitation for RAS.

To address the limitations of both deterministic and stochastic approaches, we propose the \textbf{Co}llaborative Training of \textbf{Sto}chastic and \textbf{Det}erministic Models, named CoStoDet-DDPM, where the \textit{stochastic} (DDPM) branch enhances the model's ability to handle patient-specific variations, while the \textit{deterministic} (Task) branch constrains the stochastic features of DDPM to ensure clinical relevance. Finally, the DDPM branch is discarded during inference, ensuring the real-time performance, as shown in Figure~\ref{fig:paradigms}(c). Our main contributions are:
\begin{itemize}
    \item We propose a novel dual-branch collaborative training framework that explores the simultaneous training of stochastic and deterministic models to enhance their individual performance.
    \item We introduce a new functionality of DDPM, \textit{i.e.}, feature enhancement, and applie DDPM for the first time in surgical workflow analysis. By effectively reducing prediction errors in uncertain scenarios and adapting to patient-specific anatomical and procedural variations, our approach addresses key challenge in clinical reliability.
    \item Our method achieves state-of-the-art (SOTA) results in both anticipation task (Cholec80) and recognition task (Cholec80, AutoLaparo). The inference speed is \textbf{91 FPS}, fully meeting real-time requirements.
\end{itemize}


\section{Related Works}
\label{sec:works}

\textbf{Deterministic Methods for Surgical Workflow Analysis.}
In surgical workflow analysis, nearly all methods employ deterministic approaches, which involve extracting and processing visual features from the video, and then outputting anticipation or recognition results. 
Some methods~\cite{ref20TransOpera,ref21TransTransSVNet,ref22Translovit,ref23TransCMTNet,ref24TransSkit,ref26Surgformer} use Transformer~\cite{ref6Transformer} to extract long-term features from surgical videos, but they are limited by the quadratic complexity of self-attention mechanism. 
Other research~\cite{BHI_anti22,ADA_AntiSOTA} utilize graph representation learning~\cite{GraphLearning} to model the relationships between surgical instruments and targets, enhancing model performance. 
Moreover, TMRNet~\cite{ref17lstmTMRNet} leverages video object segmentation (VOS) techniques and introduces a feature cache memory~\cite{ref28spacememory} to improve the receptive field during temporal processing. 
Ban \textit{et al}.~\cite{SuprGAN_Anti} proposed SUPR-GAN, which uses a discriminator to assess the authenticity of the anticipation, thereby enhancing the realism of the model's predictions.
BNPitfalls~\cite{ref18lstmBNPitfalls} combines CNN without batch normalization~\cite{ref29BN} with LSTM for anticipation and recognition tasks, achieving promising results.
Yang \textit{et al}.~\cite{ref25DACAT} observed that the historical information required for recognizing the current phase varies across video frames, leading to the development of DACAT, which identifies the adaptive clip for each frame and has achieved SOTA performance in surgical phase recognition. 
Additionally, \cite{ref11multi,ref12multi,ref13multimtrcnet,BHI_anti22,MIA22_Anti,ADA_AntiSOTA} utilize multi-task learning by generating extra labels, such as instrument bounding boxes or instance labels, to support the primary task (\textit{i.e.}, anticipation or recognition). 

These methods rely on deterministic feature information, making it difficult to effectively handle the significant individual variations in minimally invasive surgeries. When there is a large discrepancy between the training and test data, their performance often degrades. For instance, in anticipation task, they frequently produce false positives for instrument usage or fail to predict the timing accurately. In recognition task, they struggle to provide smooth phase recognition results or adapt to new surgical phase sequences.

\textbf{Stochastic Methods Handling Data Discrepancies.}
In domains such as robot motion planning~\cite{DiffusionPolicy,crossway}, image generation~\cite{DDPM,StableDDPM}, and image editing~\cite{PNPEdit,flexiedit,EditFT}, significant discrepancies often exist between the target and training data. Recent studies have shown that DDPM excels in these tasks. 
Chi \textit{et al}.~\cite{DiffusionPolicy} proposed Diffusion Policy, which represents the robot's visuomotor policy as a conditional denoising diffusion process to generate robot behavior. 
In real-world scenarios, even when the target environment differs significantly from training and includes human interference, the robot can still effectively achieve its goals through multimodal behavior.
In the natural domain action segmentation task, Liu \textit{et al}. proposed DiffAct~\cite{natureActSeg1} and DiffAct++~\cite{natureActSeg2}, which transform the action segmentation process into a gradual denoising process for refinement. This process introduces stochastic feature information, improving performance across multiple datasets.

The success of DDPM in the aforementioned domains demonstrates that introducing stochastic feature information helps algorithms handle data discrepancies. However, directly applying DDPM to surgical workflow analysis is insufficient due to its lack of clinical knowledge guidance and iterative denoising process. Unlike robot motion prediction, where achieving the final goal is paramount and the process itself is less critical, surgical workflows require real-time decision-making based on the specific context and anatomical details of each patient.

%% file: sec/2_methods.tex
\begin{figure}
  \centering
  \includegraphics[width=\linewidth]{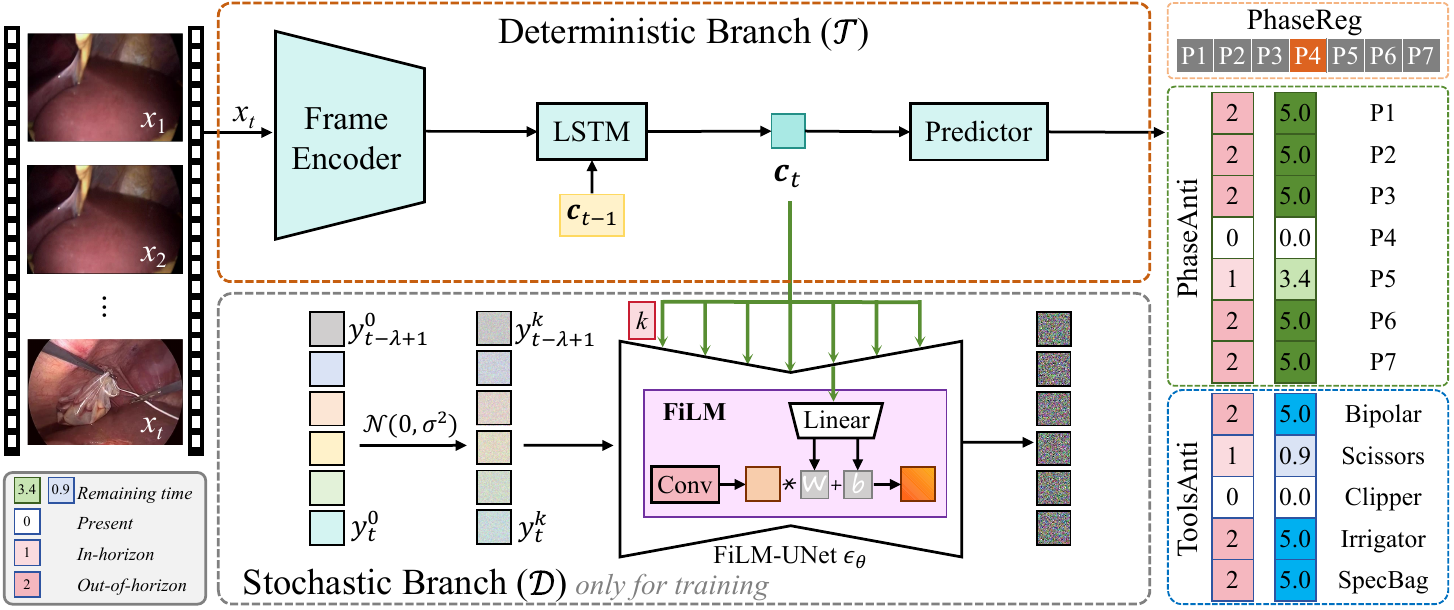}
  \caption{CoStoDet-DDPM architecture consists of a Deterministic Task Branch (\(\mathcal{T}\)) and a Stochastic DDPM Branch (\(\mathcal{D}\)). Here, \(k\) represents the time step in \(\mathcal{D}\). During training, both branches generate outputs, while during inference, only the results from \(\mathcal{T}\) are utilized to meet real-time requirements.}
  \label{fig:method}
\end{figure}

\section{Methods}
We propose CoStoDet-DDPM, a collaborative training framework integrating a Deterministic Task Branch (\(\mathcal{T}\)) and a Stochastic DDPM Branch (\(\mathcal{D}\)). 
This co-training structure ensures that $\mathcal{D}$ assists $\mathcal{T}$ in handling patient-specific variations, while $\mathcal{T}$ enforces clinical constraints on $\mathcal{D}$.
During inference, only \(\mathcal{T}\) is utilized to generate real-time outputs.

\begin{figure*}[ht]
  \centering
  \includegraphics[width=1.0\linewidth]{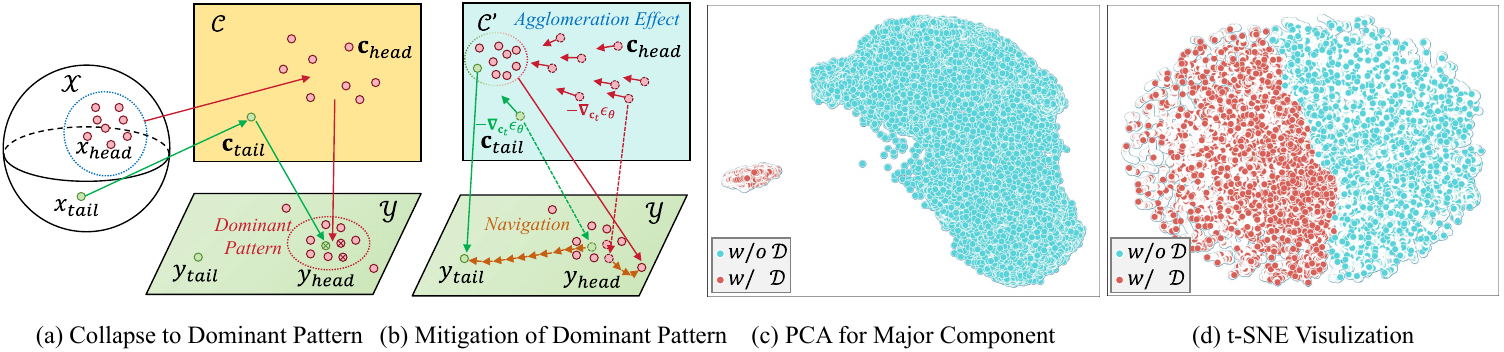}
  \caption{
    Collapse and Mitigation of the Dominant Pattern.
    In (a), without DDPM, \( \mathcal{C} \) relies solely on \( \mathcal{L}_{Task} \), leading to a sparse feature distribution. The model favors the dominant data, causing long-tail data to be misrepresented or poorly mapped.  
    After incorporating DDPM, the collapse is alleviated, as shown in (b). First, the denoising imposes an additional constraint, encouraging \( \mathbf{c}_t \) to aggregate toward regions that simultaneously satisfy both label mapping and denoising capabilities, as depicted in (c). This aggregation simplifies distribution boundary learning, facilitating global optimization.  
    Second, DDPM’s denoising process implicitly embeds ``navigation'' information into \( \mathbf{c}_t \), encoding the trajectory from noisy labels back to clean labels. This unique navigation information enables more accurate mapping and mitigates distribution disparities. As shown in (d), it is likely embedded in the minor components or local structures of \( \mathbf{c}_t \). Notably, both (c) and (d) are derived from actual feature analyses.  
  }
  \label{fig:dominantpattern}
\end{figure*}

\subsection{CoStoDet-DDPM}
We focus on two key tasks in surgical workflow analysis: anticipation and recognition. For anticipation, we follow the paradigm established in \cite{MICCAI20rethinking,MIA22_Anti} to predict the remaining time until the next surgical tool usage or phase transition. 
A time horizon \( h \) (\textit{e.g.}, 2min, 3min, or 5min) is defined as a threshold: if the remaining time exceeds \( h \), the model outputs \( h \); otherwise, it predicts the actual value. This classification distinguishes events as \textit{present}, \textit{in-horizon}, or \textit{out-of-horizon}. 
Both anticipation and recognition tasks are performed in an online manner, leveraging only historical and current information. Formally, given a video stream \( X_t = \{x_i\}_{i=1}^{t} \), where \( x_t \) is the current frame, we denote the anticipation and recognition labels as \( y^{anti}_t \) and \( y^{reg}_t \), respectively. These labels are unified as \( y_t \triangleq\{y^{anti}_t, y^{reg}_t\} \). Notably, our approach does not rely on additional location or segmentation labels.

\subsubsection{Conditional Feature Extraction}
Figure~\ref{fig:method} illustrates the architecture of CoStoDet-DDPM, which consists of three main components: (1) conditional feature extraction to obtain \(\mathbf{c}_t\), (2) the deterministic branch \(\mathcal{T}\), which utilizes \(\mathbf{c}_t\) to generate predictions, and (3) the stochastic branch \(\mathcal{D}\), which leverages \(\mathbf{c}_t\) for denoising. 
The joint operation of both branches on \(\mathbf{c}_t\) is crucial for addressing the limitations that arise when separate branches are used independently.
We employ ConvNeXt-T~\cite{ref26convnext,ref27convnextv2} as the spatial frame encoder \(\mathcal{F}_{spa}\) and an LSTM-based temporal module \(\mathcal{F}_{temp}\) for sequential processing. The spatio-temporal features \(\mathbf{c}_t \in \mathbb{R}^{512}\) are then extracted as conditional features:  
\begin{equation}  
    \mathbf{c}_{t} = \mathcal{F}_{temp}(\mathcal{F}_{spa}(x_t), \mathbf{c}_{t-1})  
\end{equation}  
where \(\mathbf{c}_{t-1}\) represents the historical features.

\subsubsection{Deterministic Branch}
In the deterministic branch \(\mathcal{T}\), we utilize $\mathbf{c}_{t}$ to directly generate final predictions through a linear predictor, with supervision provided by the task loss \(\mathcal{L}_{Task}\):  
\begin{equation}  
    \mathcal{L}_{Task} = \mathcal{L}_{reg/anti}(\mathtt{Linear}(\mathbf{c}_{t}), y_t)
    \label{eq:taskloss}
\end{equation}  

For the regression task, the standard cross-entropy loss is used, defined as:
\begin{equation}
    \mathcal{L}_{reg}=\mathcal{L}_{CE}(\hat{y}^{reg}_t,y^{reg}_t)
    \label{eq:regloss}
\end{equation}
where $\hat{y}^{reg}_t$ denotes the recognition prediction of $x_t$.

For the anticipation task, we employ the \(\mathtt{SmoothL1}\) loss~\cite{smoothl1} to supervise the prediction of the remaining time. Additionally, we introduce an auxiliary three-stage presence classification task to enhance performance, following~\cite{ref18lstmBNPitfalls,ADA_AntiSOTA}. This auxiliary task is trained using cross-entropy loss, leading to the overall loss function:  
\begin{equation}
    \begin{aligned}
        \mathcal{L}_{anti} = \mathtt{SmoothL1}(\hat{y}^{anti}_{t}, y^{anti}_t) 
        +\mu \cdot \mathcal{L}_{CE}(\hat{y}^{pres}_{t}, y^{pres}_t)
    \end{aligned}
\end{equation}  
where \(\hat{y}^{anti}_{t}\) and \(\hat{y}^{pres}_{t}\) denote the predicted remaining time and the classification output for frame \(x_t\), respectively. The trade-off factor \(\mu=0.01\) balances the auxiliary task's influence. The classification labels \(y^{pres}_t\) are derived from \(y^{anti}_t\).

\subsubsection{Stochastic Branch}
For the stochastic branch \(\mathcal{D}\), we leverage a diffusion model~\cite{DDPM,DDIM} to introduce randomness into \(\mathcal{T}\). Specifically, the features \(\mathbf{c}_t\) extracted from \(\mathcal{T}\) are used as conditioning inputs, aiming to recover the labels from a standard Gaussian distribution \(\mathcal{N}(0, I)\).
To distinguish it from the video time frame \( t \), the diffusion timestep in DDPM is denoted as \( k \). Additionally, the label \( y_t \) is represented as \( y_t^0 \).

\textit{Forward Process}. Label is corrupted by adding Gaussian noise with a predefined normalized variance schedule $ \{\beta_k\}_{k=1}^K $ at each timestep. Defining $ \bar{\alpha}_{k}=\prod_{s=1}^{k}(1-\beta_{s}) $, the noisy label $ y^k_t $ at step $ k $ is:
\begin{equation}
    y^k_t=\sqrt{\bar{\alpha}_k}y^0_t+\sqrt{1-\bar{\alpha}_k}\epsilon
\end{equation}
where $ \epsilon \sim \mathcal{N}(\textbf{0},\textbf{I}) $ and $ k \sim [1,K] $ is randomly sampled during training.

\textit{Reverse Process}. A U-Net~\cite{unet}, denoted as \( \epsilon_{\theta} \), is trained to iteratively denoise using \( \mathbf{c}_{t} \) as a condition. Unlike other DDPM-based video action segmentation methods~\cite{natureActSeg1, natureActSeg2}, which rely solely on \( \mathcal{F}_{spa}(x_t) \) as the condition and thereby lose temporal information in online manner, our approach preserves temporal dependencies by incorporating \( \mathbf{c}_{t} \).

Consequently, for both anticipation and recognition tasks, the objective is formulated as:
\begin{equation}
    \mathcal{L}_{DDPM} = \mathbb{E}_{y_t^0, \epsilon, k} \left \| \epsilon - \epsilon_{\theta}(y_t^k, k, \mathbf{c}_{t}) \right \|_2^2
    \label{eq:5_ConDDPM}
\end{equation}
During inference, a Gaussian noise map \( y^K_t \sim \mathcal{N}(\mathbf{0}, \mathbf{I}) \) is first sampled, and then denoised iteratively over \( K \) steps to recover the clean label:
\begin{equation}
    y_t^{k-1} = \frac{1}{\sqrt{\alpha_k}} \left( y_t^k - \frac{1 - \alpha_k}{\sqrt{1 - \bar{\alpha}_k}} \epsilon_{\theta}(y_t^k, k, \mathbf{c}_{t}) \right) + \sigma_k \epsilon
\end{equation}
where \( \sigma_k \) depends on \( \beta_k \).

\subsubsection{Loss Function}
To enhance clinical consistency, we extend the label \( y_t^0 \) backward by \( \lambda = 32 \) historical frames for \( \mathcal{D} \), allowing DDPM to better capture the temporal context and clinical variations.
The overall loss for CoStoDet-DDPM is:
\begin{equation}
    \mathcal{L} = \mathcal{L}_{Task} + \mathcal{L}_{DDPM}
\end{equation}
This enables the model to learn both task-specific predictions and enhanced feature representations by leveraging DDPM's stochastic properties. During inference, only \( \mathcal{T} \) is retained, ensuring real-time application.

\subsection{Uncertainty-Aware Feature Enhancement} 

When solely optimizing the training objective in Eq.~\ref{eq:taskloss}, a major issue arises: \textbf{Collapse to Dominant Patterns}.  
The introduction of the Markov diffusion chain \( \mathcal{D} \) helps mitigate this by enhancing feature uncertainty awareness in \( \mathcal{T} \), enabling it to better capture patient-specific variations and improving model robustness.  
The primary change introduced by \( \mathcal{D} \) lies in the transformation of \( \mathbf{c}_t \) within the feature space \( \mathcal{C} \), ensuring more diverse and informative representations.
To distinguish between them, we denote the features and feature space of pure \(\mathcal{T}\) as \( \mathbf{c}_t \) and \( \mathcal{C} \), respectively, while those incorporating DDPM are denoted as \( \mathbf{c}'_t \) and \( \mathcal{C}' \).

\textbf{(1) The Agglomeration Effect.}  
The original conditional feature space \( \mathcal{C} \subset \mathbb{R}^{512} \). When optimized solely using \( \mathcal{L}_{Task} \), \( \{\mathbf{c}_t\} \subset \mathcal{C} \) are widely scattered. This results in the linear mapping \( \mathtt{Linear} \triangleq f_\phi \) being dominated by high-frequency data, causing the model to underfit long-tail data and leading to pattern collapse, as shown in Figure~\ref{fig:dominantpattern}(a).

To address this, we introduce an optimization term \( \mathcal{L}_{DDPM} \), which enforces a new constraint on the formation of the feature space \( \mathcal{C}' \). In \( \mathbb{R}^{512} \), only a small subset of the space can simultaneously satisfy both the label mapping and denoising requirements. This constraint pulls \( \mathbf{c}_t \) towards a new position \( \mathbf{c}'_t \) endowed with robust denoising capabilities:
\begin{equation}
    \Delta \mathbf{c}_t \propto -\eta \left( |f_{\phi} - y_t| \nabla_{\mathbf{c}_t} f_{\phi}(\mathbf{c}_t) + |\epsilon_{\theta} - \epsilon| \nabla_{\mathbf{c}_t} \epsilon_{\theta}(y_{t}^k, k, \mathbf{c}_t) \right)
    \label{eq:delta-DDPM}
\end{equation}
where \( \eta \) represents the learning rate.  
As a result, incorporating \( \mathcal{L}_{DDPM} \) compacts the feature distribution in \( \mathcal{C}' \), preventing the model from over-relying on dominant patterns. The visualization in Figure~\ref{fig:dominantpattern}(c) further substantiates this effect.

The benefit of this aggregated feature space is twofold. First, it simplifies the learning of distribution boundaries, making it easier for the model to achieve global optimality rather than settling for local optima driven by dominant data. Second, the implicit regularization imposed by DDPM forces the linear layer to classify based on intrinsic semantic features rather than mere statistical biases, thus preventing the dominant pattern from obscuring long-tail data.




\textbf{(2) Embedding of Navigation Information.}
During the learning process of \( \mathbf{c}'_t \) with the assistance of \( \mathcal{L}_{DDPM} \), \( \mathbf{c}'_t \) must satisfy the constraints of the noise path: $\forall k \in[1,K]$, $ \epsilon_{\theta}(y_{t}^k, k, \mathbf{c}'_t) \approx \epsilon$. This requirement inherently embeds the ``navigation'' path from \( y_t^k \) back to \( y_t^0 \) within \( \mathbf{c}'_t \).  

The mapping from \( \mathbf{c}'_t \) to target is essentially equivalent to following a predefined roadmap to accurately locate the correct target. For both dominant and long-tail data, each \( x_t \) is associated with a distinct \( \mathbf{c}'_t \) that embeds unique navigation information (\textit{i.e.}, a map), enabling more precise target identification and mitigating the collapse of dominant patterns. These navigation cues may be encoded in the minor components or local structures of \( \mathbf{c}'_t \), preserving the agglomeration of its primary components, as illustrated in Figure~\ref{fig:dominantpattern}(d).
Combining (1) and (2), the dominant pattern collapse is mitigated, as shown in Figure~\ref{fig:dominantpattern}(b).

\subsection{Clinical-Semantic Constraint}

In CoStoDet-DDPM, \(\mathcal{T}\) plays a crucial role in assisting \(\mathcal{D}\) to maintain clinical consistency.  

However, in pure DDPM, the conditioning signal tends to weaken during model training and noise addition, which is undesirable for surgical workflow analysis. To counteract this, the task loss term \(\mathcal{L}_{Task}\) (Eq.~\ref{eq:taskloss}) reinforces the conditioning effect. As demonstrated in \cite{DDPMTheory}, minimizing the objective in Eq.~\ref{eq:5_ConDDPM} for conditional DDPM is equivalent to:  
\begin{equation}
    \mathcal{L}_{DDPM} \sim \mathbb {E}_{y_{t}^0,k} \left \|y_{t}^0 -y _{\theta} \left ( y_t^k,k,\mathbf{c}_{t}\right )   \right \|^{2}_2
    \label{eq:6_ConTDDPM}
\end{equation}  
where \( y_\theta \) denotes the network that directly reconstructs the clean labels from the noisy input.  

The task loss \(\mathcal{L}_{Task}\) not only enforces a semantic alignment between \(\mathbf{c}_{t}\) and \(y_t\) but also guides DDPM’s denoising trajectory, ultimately facilitating the minimization of Eq.~\ref{eq:6_ConTDDPM}.
We infer that in task-specific conditional DDPMs, aligning the conditioning input with the task label ensures that the features extracted by \(\mathcal{F}_{spa}\) and \(\mathcal{F}_{temp}\) remain highly relevant to the given task. Even if certain information is lost during feature extraction, as long as the extracted features can be directly mapped to the labels, DDPM retains sufficient information to recover the labels from noise.


%% file: sec/3_experiment.tex
\section{Experiments}
\subsection{Experimental Design}

\textbf{Datasets.}
We conducted experiments on two public surgical datasets: Cholec80~\cite{twinanda2016endonet} and AutoLaparo~\cite{wang2022autolaparo}. \textit{Cholec80} comprises 80 laparoscopic cholecystectomy videos, with durations ranging from 15 to 90 minutes at 25 FPS. Following previous works, Cholec80 is split into 60 training and 20 testing for the anticipation task~\cite{MIA22_Anti,ADA_AntiSOTA}, and 40 training and 40 testing for the recognition task~\cite{ref24TransSkit,ref25DACAT,ref26Surgformer}. 
\textit{AutoLaparo} contains 21 laparoscopic hysterectomy videos with an average duration of 66 minutes recorded at 25 FPS. This dataset is divided into 10 training, 4 validation, and 7 testing videos following~\cite{ref24TransSkit,ref25DACAT,ref26Surgformer} for recognition task. Both datasets are resampled into 1 FPS as~\cite{ref18lstmBNPitfalls,ref25DACAT,MIA22_Anti,ADA_AntiSOTA}.

\textbf{Metrics.}
For the anticipation task, we use frame-based metrics, including mean absolute error (MAE) and its variants: \(eMAE\), \(outMAE\), and \(wMAE\), across three time horizons $h$: 2min, 3min, and 5min. \(eMAE\) measures the anticipation error for events occurring within $0.1h$ in the future, while \(outMAE\) evaluates \textit{out-of-horizon} events. \(wMAE\) provides an average error measurement for \textit{in-horizon} and \textit{out-of-horizon} events.
Due to factors such as surgical conditions and surgeon autonomy, the progression of surgical events may accelerate or decelerate, and the actual remaining time is not always linear. However, within a $0.1h$ window, events approximately follow a linear progression, making \(eMAE\) the most important metric. \(outMAE\) measures whether the model produces false positive anticipations. 
For the recognition task, we employ four widely-used metrics, \textit{i.e.}, accuracy, precision, recall, and Jaccard. 

\textbf{Inplementation Details.}
All experiments were conducted on a single NVIDIA GeForce RTX 4090 24GB GPU, with separate models trained for the anticipation and recognition tasks. 
For the anticipation task, we trained for 300 epochs using the AdamW optimizer with a learning rate of $1e-4$, weight decay of $1e-6$, batch size of 1. In each training iteration, a randomly selected 64-frame clip was extracted from a single video.
For the recognition task, we trained for 30 epochs using AdamW optimizer with a learning rate of $1e-5$, weight decay of $1e-2$, and batch size of 1. During training, non-overlapping 64-frame clips wextracted sequentially from a single video.
For both tasks, we used a cosine warm-up schedule, and the number of DDPM timesteps was set to 100.

\subsection{Comparison With State-of-the-arts}
\textbf{Anticipation.} We compare our approach with recent SOTA methods on Cholec80, and the results are listed in Table~\ref{tab:tab1_sota_anti}. Each horizon was trained three times, as done in other works. The results show a significant improvement with our approach, particularly at $h=5$min.  
For the $eMAE$ metric, the relative improvements are 16.1\% and 11.1\% for tool and phase predictions, respectively. A lower $eMAE$ indicates that our method provides more accurate remaining time anticipation, and the lower $outMAE$ suggests a reduced probability of generating false positive alerts. These strong performance gains highlight the effectiveness of our method and underscore the importance of combining both deterministic and stochastic models for surgical anticipation.

\definecolor{SilverChalice}{rgb}{0.65,0.65,0.65}
\begin{table*}
\centering
\resizebox{0.8\linewidth}{!}{
\begin{tblr}{
  cells = {c},
  row{7} = {fg=SilverChalice},
  row{8} = {fg=SilverChalice},
  row{9} = {fg=SilverChalice},
  cell{1}{1} = {r=3}{},
  cell{1}{2} = {c=6}{},
  cell{2}{2} = {c=3}{},
  cell{2}{5} = {c=3}{},
  vline{2} = {1-3}{},
  vline{5} = {2-3}{},
  vline{2,5} = {4-12}{},
  hline{1,4,11-12} = {-}{},
  hline{2-3} = {2-7}{},
  hline{1,12} = 1pt
}
Methods     & $wMAE$~$\downarrow$  / $outMAE$~$\downarrow$ / $eMAE$~$\downarrow$ & & & & & \\
           & Tool& & & Phase & & \\
           & $h=2$ min & $h=3$ min & $h=5$ min & $h=2$ min & $h=3$ min & $h=5$ min \\
TimeLSTM~\cite{TimeLSTM_anti_miccai17}   & $0.51/0.16/1.56$ & $0.76/0.29/2.26$ & $1.32/0.67/3.62$ & $0.47/0.22/1.29$ & $0.64/0.32/1.73$                           & $1.07/0.60/2.60$                         \\
RSDNet~\cite{rsdnet_anti_TMI18}     & $0.48/0.23/1.25$                         & $0.73/0.42/1.80$                 & $1.26/0.90/2.61$                         & $0.43/0.20/1.15$                         & $0.63/0.33/1.55$                           & $1.10/0.65/2.40$                         \\
Bayesian~\cite{MICCAI20rethinking}   & $0.43/0.08/1.12$                         & $0.66/0.15/1.65$                 & $1.09/0.44/2.68$                         & $0.39/0.15/1.02$                         & \uline{$0.59$}$/0.32/1.47$                           & $0.85/0.52/1.54$                         \\
IIA-Net~\cite{MIA22_Anti}    & $0.38/0.10/1.01$                         & $0.58/0.19/1.46$                 & $0.92/0.44/2.14$                         & $0.36/0.10/1.18$                         & $0.49/0.18/1.42$                           & $0.68/0.28/1.09$                         \\
GCN-MSTCN~\cite{BHI_anti22}  & $0.48/0.33/0.87$                         & $0.72/0.52/1.26$                 & $1.21/0.99/1.90$                         & $0.45/0.29/0.77$                         & $0.67/0.47/1.06$                           & $1.06/0.85/1.46$                         \\
ADANet~\cite{ADA_AntiSOTA}     & $0.39/0.07/0.99$                         & $0.57/0.19/1.21$                 & $1.03/0.73/1.49$                         & $0.35/0.07/0.95$                         & $0.47/0.17/1.03$                           & $0.80/0.55/1.06$                         \\
BNPitfalls~\cite{ref18lstmBNPitfalls} & \uline{$0.40$}$/$\uline{$0.06$}$/$\uline{$1.06$} & \uline{$0.58$}$/$\uline{$0.13$}$/$\uline{$1.42$} & \uline{$0.96$}$/$\uline{$0.41$}$/$\uline{$2.11$} & \uline{$0.29$}$/$\uline{$0.06$}$/$\uline{$0.63$} & $\mathbf{0.39}$$/$\uline{$0.10$}$/$\uline{$0.78$}          & \uline{$0.61$}$/$\uline{$0.24$}$/$\uline{$0.99$} \\
Ours       & $\mathbf{0.39}/\mathbf{0.05}/\mathbf{0.93}$       & $\mathbf{0.56}/\mathbf{0.09}/\mathbf{1.28}$        & $\mathbf{0.91}/\mathbf{0.29}/\mathbf{1.77}$                & $\mathbf{0.27}/\mathbf{0.05}/\mathbf{0.52}$                & $\mathbf{0.39}/\mathbf{0.09}/\mathbf{0.67}$ & $\mathbf{0.59}/\mathbf{0.18}/\mathbf{0.88}$                
\end{tblr}
}
\caption{Comparison results (min) with SOTA on Cholec80. Methods in \textcolor{SilverChalice}{gray} are not suitable for direct comparision due to the extra usage of dataset and annotation. ADANet~\cite{ADA_AntiSOTA} only considers anticipation for six phases. \textbf{Bold}: Best result. \underline{Underline}: Second-best result.}
\label{tab:tab1_sota_anti}
\end{table*}

\begin{figure}[t]
\centerline{\includegraphics[width=\columnwidth]{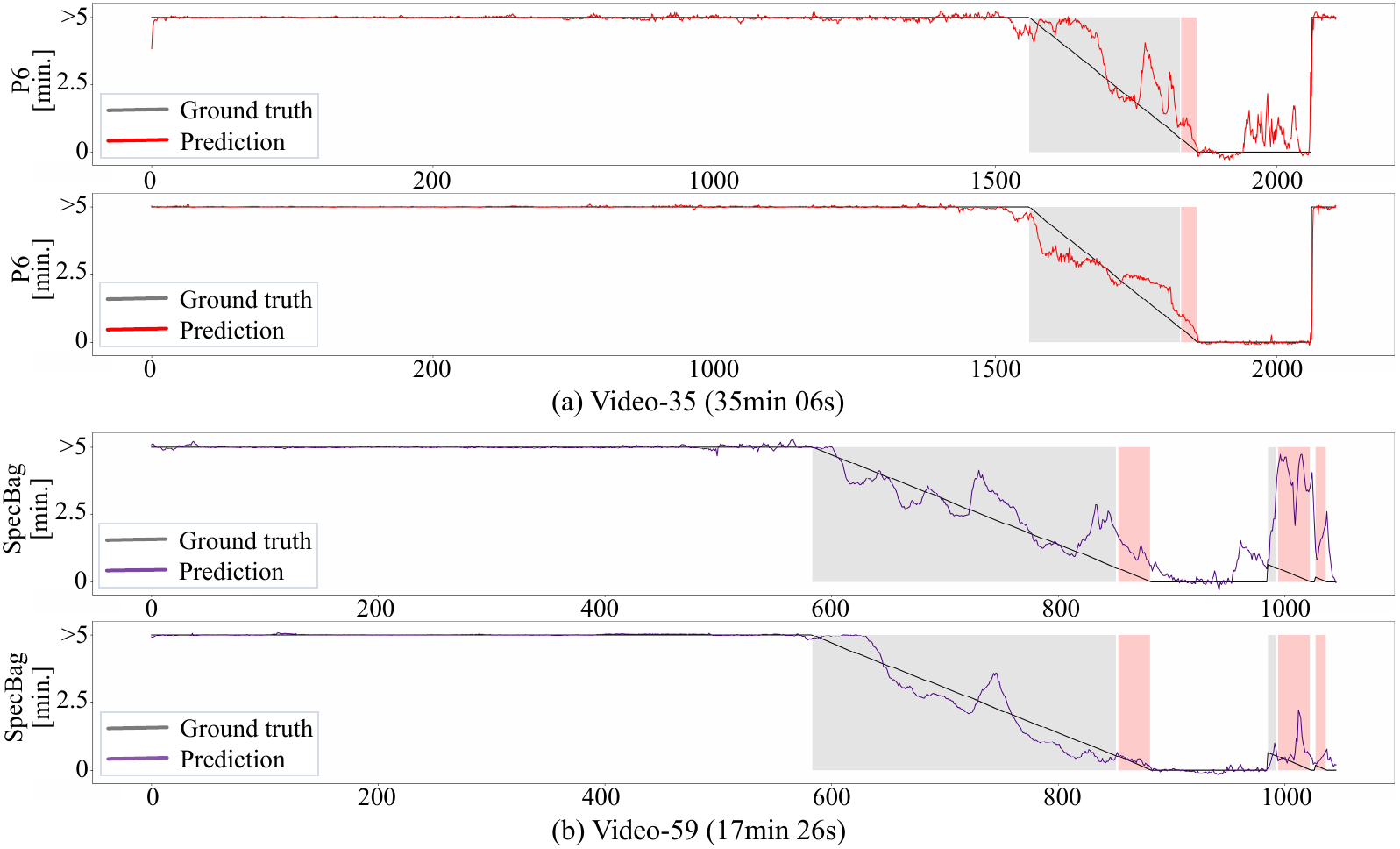}}
\caption{Visualization comparison for the anticipation task. (a) shows the phase, and (b) displays the tool visualization. In each sub-figure, the top row represents the SOTA method BNPitfalls, and the bottom row represents ours.} 
\label{fig:fig3_visual_anti}
\end{figure}

Figure~\ref{fig:fig3_visual_anti} illustrates the visualizations for phase and tool anticipation. It is evident that our CoStoDet-DDPM outperforms in the gray areas, with especially strong performance in the red regions, which correspond to the $eMAE$ metric. Additionally, our model exhibits smoother predictions for both \textit{out-of-horizon} and \textit{present} events, emphasizing its improved accuracy and stability.

\begin{figure}[t]
\centerline{\includegraphics[width=\columnwidth]{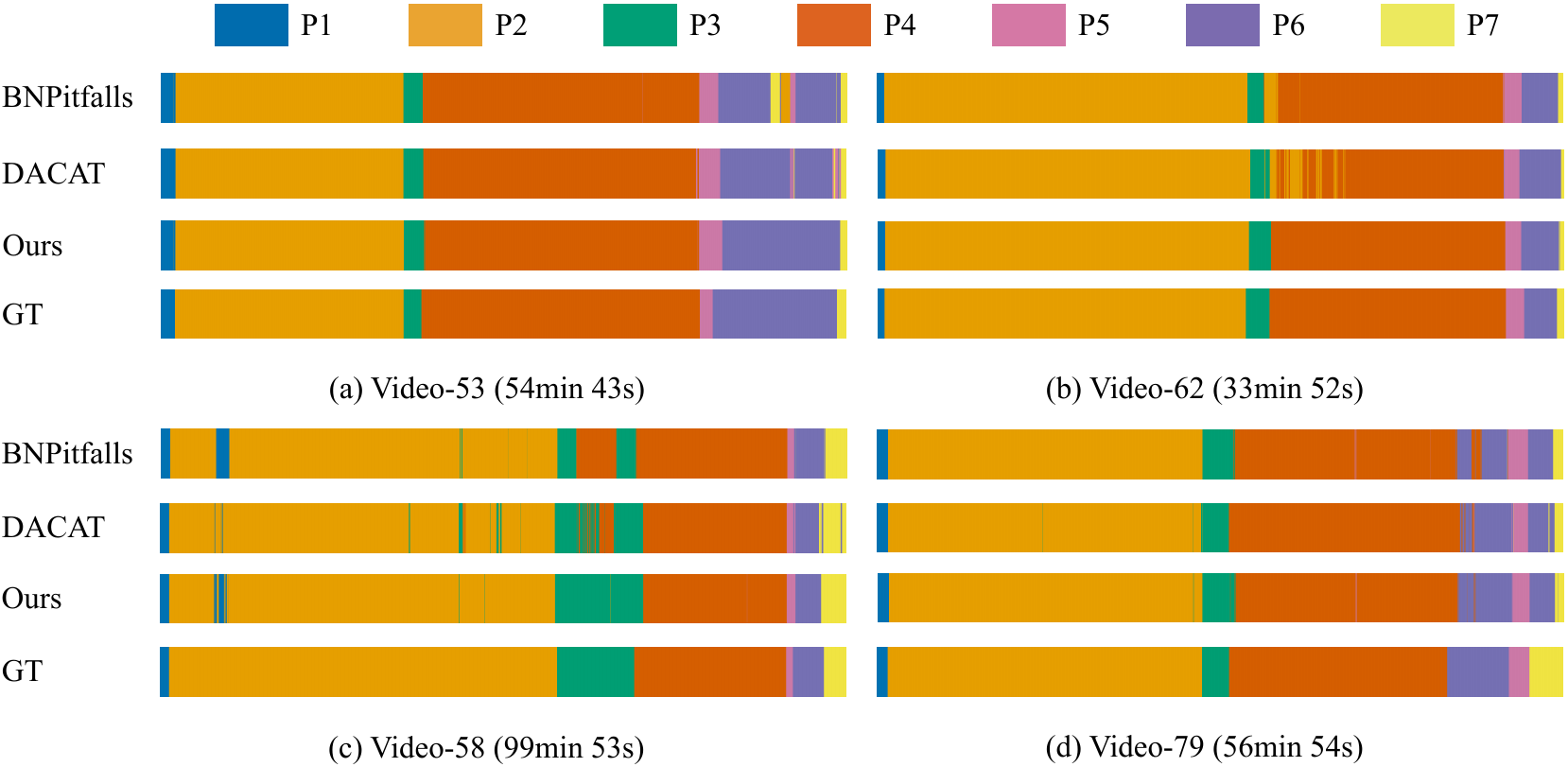}}
\caption{Visualization comparison for the recognition task. (a) and (b) showcase accurate predictions, while (c) and (d) depict relatively poorer results.} 
\label{fig:fig4_visual_rec}
\end{figure}

\textbf{Recognition.} We compare ours with previous SOTAs on Cholec80 and AutoLaparo, the results are shown in Table~\ref{tab:tab2_sota_reg}. 
Our method achieves relative improvements of 1.0\% and 1.5\% in the Jaccard index on the two datasets, respectively. Combined with the visualization results in Figure~\ref{fig:fig4_visual_rec}, it is evident that ours produces smoother results within phases and more accurate phase transitions. Both the metrics and visualization results show that after incorporating the stochastic model, our approach alleviates the issues of collapse to dominant patterns that are present in pure deterministic models. Overall, our method demonstrates strong capability in handling patient-specific variations for both anticipation and recognition tasks in surgical workflow analysis.

\begin{table}[t]
    \centering
    \resizebox{\linewidth}{!}{
    \begin{tabular}{c|c|cccc} \specialrule{1pt}{0pt}{0pt}
    Datasets & Methods & Accuracy $\uparrow$ & Precision $\uparrow$ & Recall $\uparrow$ & Jaccard $\uparrow$ \\ \hline
    \multirow{9}{*}{Cholec80} 
    & Trans-SVNet~\cite{ref21TransTransSVNet} & $89.1 \pm 7.0$ & $84.7$ & $83.6$ & $72.5$ \\
    & AVT~\cite{girdhar2021anticipative} & $86.7 \pm 7.6$ & $77.3$ & $82.1$ & $66.4$ \\
    & TeSTra~\cite{zhao2022real} & $90.1 \pm 6.6$ & $82.8$ & $83.8$ & $71.6$ \\
    & LoViT~\cite{ref22Translovit} & $91.5 \pm 6.1$ & $83.1$ & $86.5$ & $74.2$ \\
    & SKiT~\cite{ref24TransSkit} & $92.5 \pm 5.1$ & 84.6 & 88.5 & 76.7 \\
    & Surgformer~\cite{ref26Surgformer} & $92.4 \pm 6.4$ & $87.9$ & $\underline{89.3}$ & $79.9$ \\
    & BNPitfalls~\cite{ref18lstmBNPitfalls}& $93.7 \pm 4.1$&\underline{$88.7$} & $87.7$ & $79.2$  \\
    & DACAT~\cite{ref25DACAT} & $\underline{94.1 \pm 4.3}$ & $\mathbf{89.2}$ & $88.5$ & $\underline{80.6}$ \\
    & Ours & $\mathbf{94.5 \pm 3.6}$ & $88.4$ & $\mathbf{90.2}$ & $\mathbf{81.4}$ \\
    \hline

    \multirow{10}{*}{AutoLaparo} 
    & SV-RCNet~\cite{ref14lstmSV-RCNet} & $75.6$ & $64.0$ & $59.7$ & $47.2$ \\
    & TMRNet~\cite{ref17lstmTMRNet} & $78.2$ & $66.0$ & $61.5$ & $49.6$ \\
    & Trans-SVNet~\cite{ref21TransTransSVNet} & $78.3$ & $64.2$ & $62.1$ & $50.7$ \\
    & LoViT~\cite{ref22Translovit} & $81.4 \pm 7.6$ & $\mathbf{85.1}$ & $65.9$ & $55.9$ \\
    & SKiT~\cite{ref24TransSkit} & $82.9 \pm 6.8$ & $81.8$ & $70.1$ & $59.9$ \\
    & SPHASE~\cite{ref33sphase} & $83.8$ & $75.7$ & $71.3$ & $57.8$ \\
    & Surgformer~\cite{ref26Surgformer} & $85.7\pm 6.9$ & $82.3$ & $\underline{75.7}$ & $66.7$ \\
    & BNPitfalls~\cite{ref18lstmBNPitfalls} & $86.8\pm1.5$ & $78.2$ & $72.0$ & $64.2$ \\
    & DACAT~\cite{ref25DACAT} & $\mathbf{87.8 \pm 7.6}$ & $78.5$ & $75.0$ & $\underline{66.9}$ \\
    & Ours & $\underline{87.6\pm7.8}$ & $\underline{84.5}$ & $\mathbf{76.4}$ & $\mathbf{67.9}$ \\
    \hline
    \specialrule{1pt}{0pt}{0pt}
    \end{tabular}
    }
\caption{Comparison results (\%) with SOTA on Cholec80 and AutoLaparo. \textbf{Bold}: Best result. \underline{Underline}: Second-best result.}
\label{tab:tab2_sota_reg}
\end{table}

\subsection{Ablation Study}
\textbf{The Role of $\mathcal{D}\rightarrow \mathcal{T}$ and $\mathcal{T}\rightarrow \mathcal{D}$.}
We conducted ablation experiments on the Cholec80 using the anticipation task. Specifically, we investigate the impact of $\mathcal{D} \rightarrow \mathcal{T}$ and $\mathcal{T} \rightarrow \mathcal{D}$. To better evaluate the volatility of predictions from different branches, we propose a new metric, $Smooth$, defined as:
\begin{equation}
    Smooth=\frac{1}{|N|-1} \sum_{i=1}^{|N|-1}\left|N_{i+1}-N_i\right|
\end{equation}
where $N = \left\{ p_t^{anti} \mid y_t^{anti} = h, t \in \mathbb{T} \right\}$, and $\mathbb{T}$ represents the set of predictions in the \textit{out-of-horizon} region and the time length of the video, respectively.

\begin{figure}[t]
\centerline{\includegraphics[width=\columnwidth]{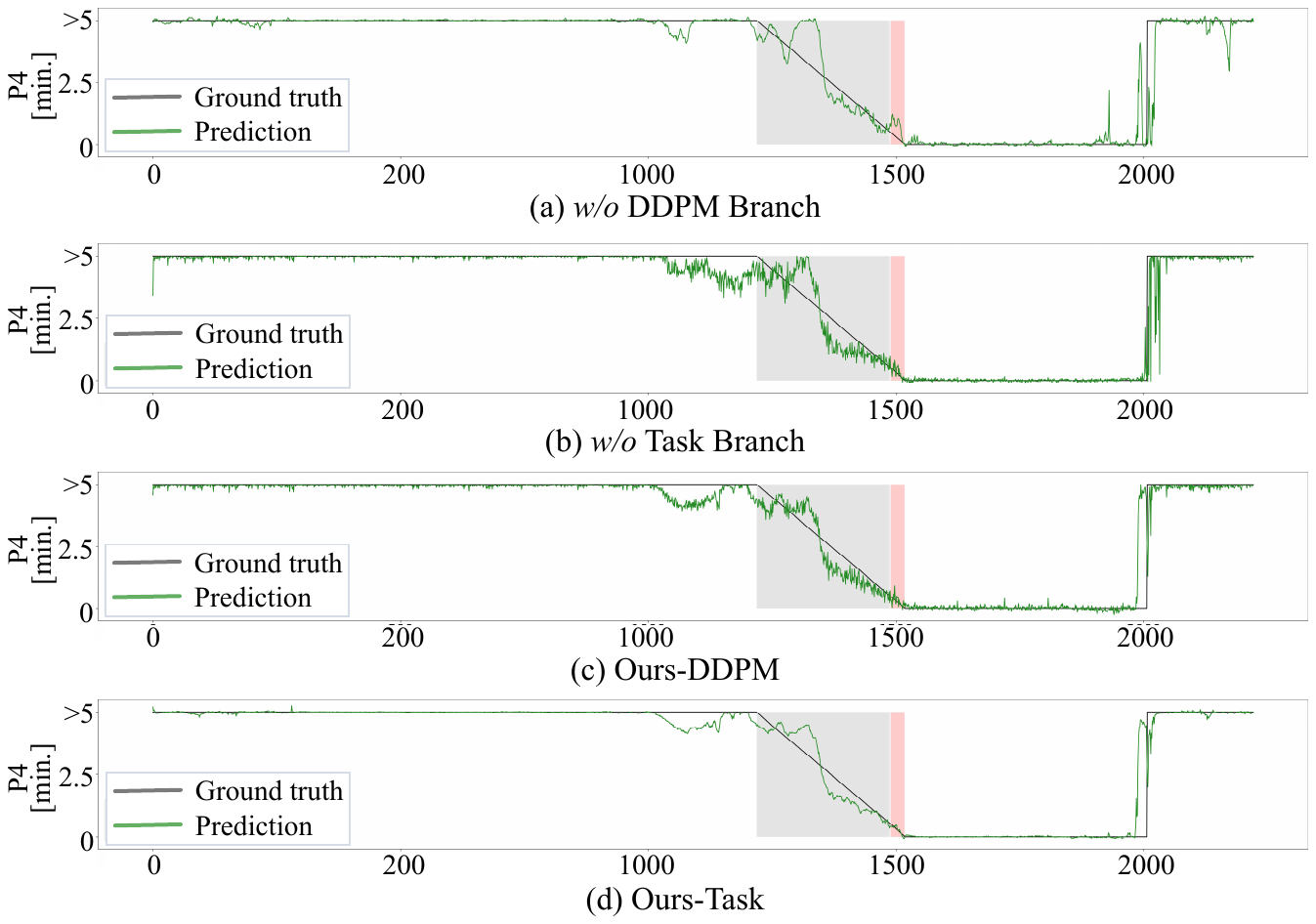}}
\caption{Anticipations of different types of models. (a) and (b) show the pure deterministic and pure stochastic models, respectively. (c) and (d) present the our results from $\mathcal{D}$ and $\mathcal{T}$, respectively. (b) and (d) use DDIM for 16 steps of inference.} 
\label{fig:fig5_aba1}
\end{figure}

\begin{table}[t]
\centering
\resizebox{\linewidth}{!}{
\begin{tabular}{c|ccc} 
\specialrule{1pt}{0pt}{0pt}
Methods                  & Tool                    & Phase                   & $Smooth$                \\ 
\hline
$w/o$~DDPM Branch & $0.96/0.41/2.11$          & $0.61/0.24/0.99$          & $0.053+0.042$           \\
$w/o$~Task Branch & $0.97/0.34/2.03$          & $0.63/0.19/1.05$          & $0.133+0.079$           \\ 
\hline
Ours-DDIM                & $0.93/0.34/\mathbf{1.75}$ & $0.62/0.23/\mathbf{0.86}$ & $0.095+0.077$           \\
Ours-Task                & $\mathbf{0.91}/\mathbf{0.29}/1.77$ & $\mathbf{0.59}/\mathbf{0.18}/0.88$ & $\mathbf{0.025}+\mathbf{0.020}$  \\
\specialrule{1pt}{0pt}{0pt}
\end{tabular}
}
\caption{The results of different types of models with $h=5$ min. The columns showcast the $wMAE/outMAE/eMAE$, while $Smooth$ are reported as $Smooth_{Tool}+Smooth_{Phase}$.}
\label{tab:tab3_abla}
\end{table}

As shown in Table~\ref{tab:tab3_abla}, it is evident that without the stochastic model, the task branch fails to provide accurate predictions. Conversely, without the deterministic model, DDPM lacks clinical consistency and is prone to false positives. Our collaborative learning framework overcomes the limitations in each branch, leading to results from both branches surpassing the SOTA performance. The visualization results in Figure~\ref{fig:fig5_aba1} further highlight how $\mathcal{D}$ and $\mathcal{T}$ complement each other, particularly in uncertain regions, such as the red areas.



\textbf{Different Collaborative Learning Strategies.}
\label{exp:co-training}
From the perspective of $\mathcal{T}$, $\mathcal{D}$ effectively enhances features through collaborative learning. In this ablation study, we build upon the SOTA BNPitfalls (BNP) and investigate the impact of different collaborative learning strategies on its performance. The strategies can be broadly categorized into two groups: pre-trained methods and co-training methods. The former includes Masked Autoencoders (MaskAE)~\cite{he2022masked}, while the latter encompasses CURL~\cite{laskin2020curl}, GC~\cite{wang2024groupcontrast}, and RandomMask.
CURL enhances the feature extraction and processing ability of model by maximizing the similarity between features extracted from the original observation and those extracted after data augmentation (\textit{e.g.}, random cropping).
GC combines semantic-aware contrastive learning with segment grouping to improve feature representation by leveraging semantic coherence and reducing semantic conflicts.
Inspired by the derivation process of our method, we designed RandomMask to replace the DDPM denoising task with the reconstruction of randomly masked labels using the same conditional features. FiLM-UNet is then used to restore the original features, introducing an additional task for co-training.

\begin{figure}[t]
\centerline{\includegraphics[width=\columnwidth]{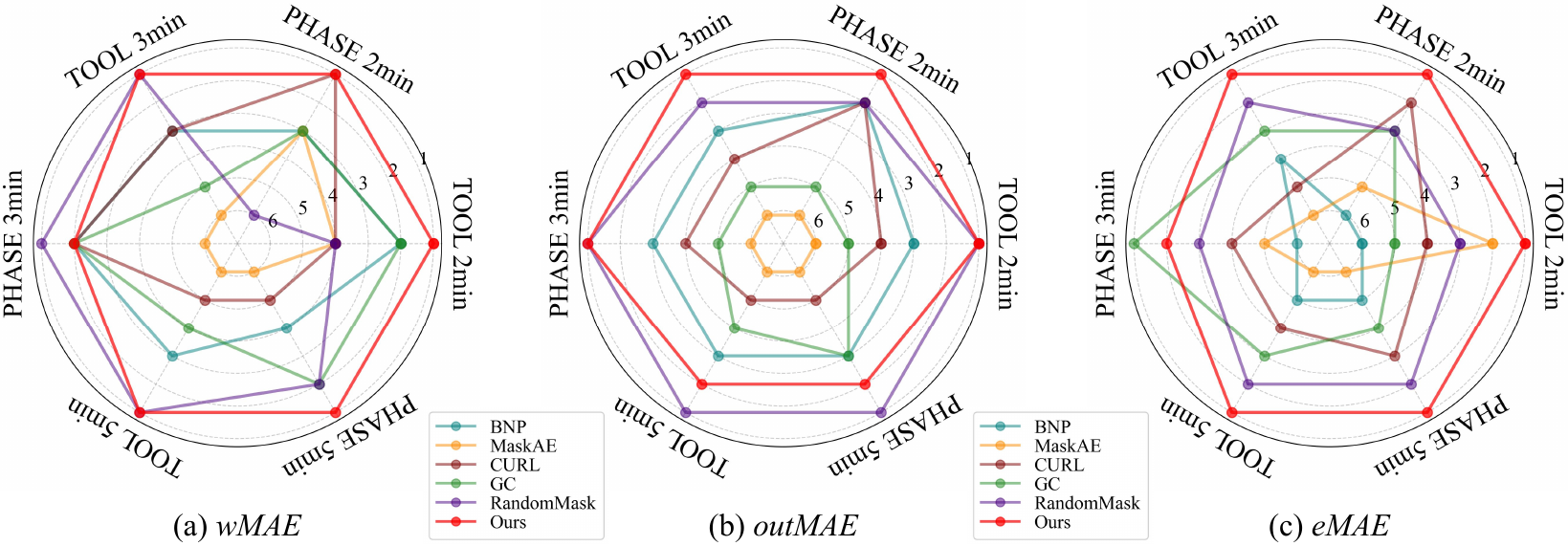}}
\caption{The impact of co-training methods, including MaskAE~\cite{he2022masked}, CURL~\cite{laskin2020curl}, GC~\cite{wang2024groupcontrast}, RandomMask and Ours, all built upon the BNPitfalls (BNP). The scale represents rankings, with detailed data provided in the Supplementary Materials.} 
\label{fig:fig7_aba3}
\end{figure}

The experimental results are shown in Figure~\ref{fig:fig7_aba3}. Overall, GC, RandomMask, and our $\mathcal{D}$ all provide certain improvements over BNPitfalls, with the largest improvement coming from $\mathcal{D}$. MaskAE, as a pre-training method, performs poorly, possibly due to the limited size of the surgical dataset Cholec80, which hinders its effectiveness. CURL, on the other hand, struggles to handle patient-specific variations due to its co-training approach. In summary, $\mathcal{D}$, which does not directly modify the features but instead assigns denoising functionality and introduces diversity through noise in the co-training process, can enhance the performance of existing methods in anticipation task.

\begin{figure}[t]
\centerline{\includegraphics[width=\columnwidth]{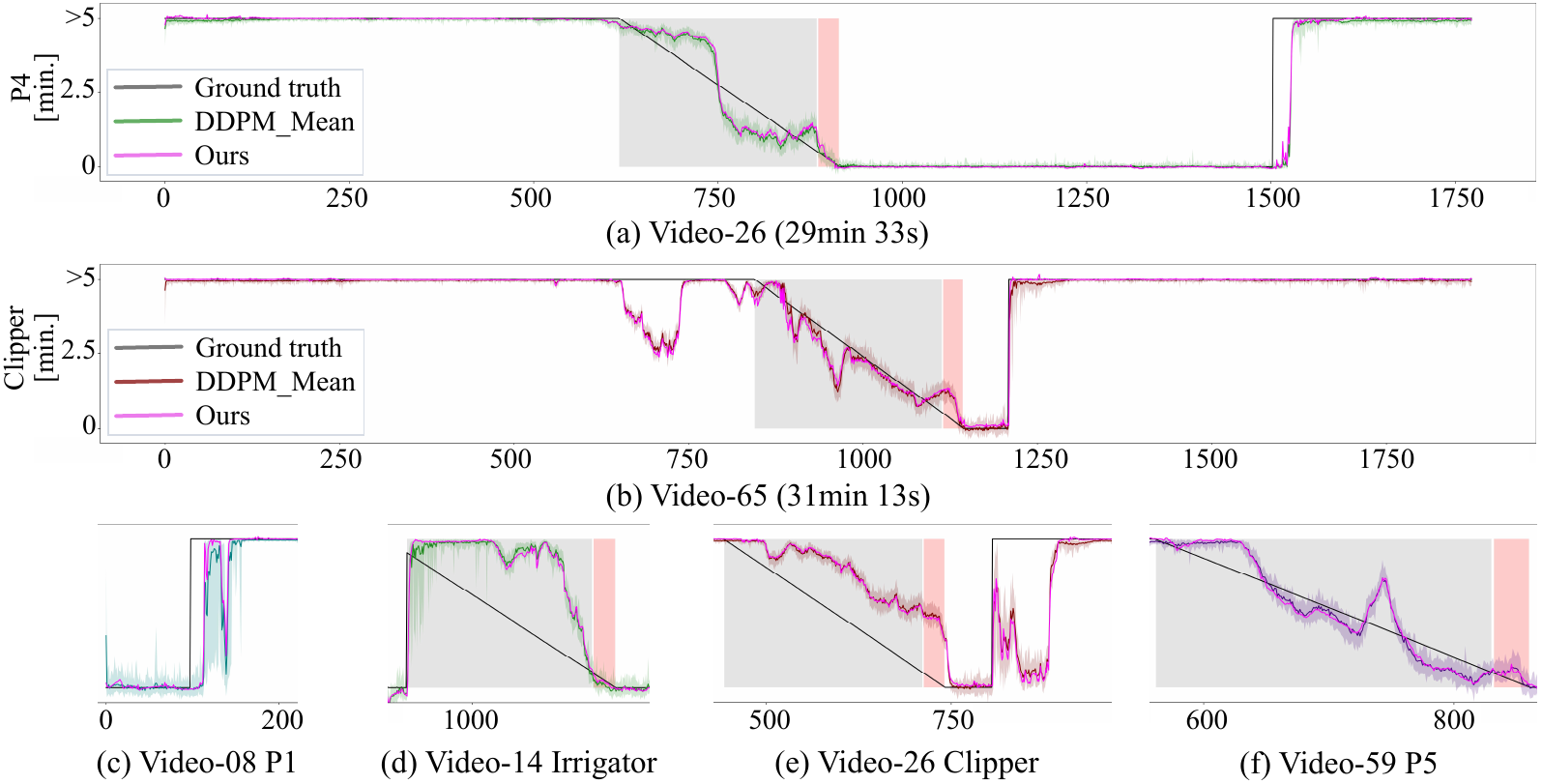}}
\caption{The output relationship between $\mathcal{T}$ and $\mathcal{D}$.
The pink solid line represents the predictions of $\mathcal{T}$, while the other solid lines indicate the mean predictions of $\mathcal{D}$ using 20 random seeds during inference. The shaded area are the range of predictions from $\mathcal{D}$ across the 20 seeds.} 
\label{fig:fig8_aba4}
\end{figure}

\textbf{The Output Relationship Between $\mathcal{T}$ and $\mathcal{D}$.}
Disregarding inference speed, we investigate the relationship between the outputs of the two branches, $\mathcal{T}$ and $\mathcal{D}$, of CoStoDet-DDPM from a statistical perspective. For $\mathcal{D}$, we randomly select 20 different seeds and perform inference using DDIM with 16 steps. We then plot the value range formed by the 20 different curves along with their mean values, as shown in Figure~\ref{fig:fig8_aba4}. In the figure, the pink solid line represents the output of $\mathcal{T}$, while the other colored lines represent the mean (or expectation) output of $\mathcal{D}$. The shaded area indicates the output range of $\mathcal{D}$.

From the figure, it can be observed that the pink solid line closely aligns with the other colored lines, indicating that the output of $\mathcal{T}$ closely approximates fitting the output expectation of $\mathcal{D}$ across different seeds. This is the result of the co-training process. When the output of $\mathcal{T}$ is uncertain, such as in the \textit{in-horizon} regions of (a), (b), (d), (e), and (f) and the\textit{ out-of-horizon} regions of (c) and (e), the output of $\mathcal{D}$ shows greater fluctuations. This clearly demonstrates that the two branches underwent deep coupled learning during training, each contributing its strengths. Ultimately, this co-learning process enables $\mathcal{T}$ to produce better results.

\textbf{$\mathcal{D}$ as a Feature Enhancer for Improving SOTA.}
We further explore the improvement brought by $\mathcal{D}$ as a feature enhancer for SOTA methods. For the anticipation task, the performance improvement is shown in the last two rows of Table~\ref{tab:tab1_sota_anti}, where $\mathcal{D}$ significantly reduces the error of BNPitfalls. For the recognition task, the improvement is demonstrated in Table~\ref{tab:tab5_abla_5_2}. We applied $\mathcal{D}$ to two recent SOTA methods, namely BNPitfalls and DACAT, and achieved improvements in both cases. This demonstrates that DDPM, as a feature enhancer, can improve the accuracy of surgical workflow recognition tasks.

\begin{table}[t]
\centering
\resizebox{\linewidth}{!}{
\begin{tblr}{
  cells = {c},
  cell{2}{1} = {r=4}{},
  cell{6}{1} = {r=4}{},
  vline{2-3} = {1-9}{},
  hline{1-2,6,10} = {-}{},
  hline{4,8} = {2-6}{dashed},
  hline{1,10} = 1pt
}
Datasets   & Methods & Accuracy~$\uparrow$  & Precision~$\uparrow$  & Recall~$\uparrow$   & Jaccard~$\uparrow$   \\
Cholec80   & BNPitfalls & $93.7\pm4.1$          & $88.7$          & $87.7$          & $79.2$          \\
           & Ours       & $\mathbf{94.1\pm3.5}_{\color{red}+0.4}$ & $\mathbf{88.9}_{\color{red}+0.2}$ & $\mathbf{89.2}_{\color{red}+1.5}$ & $\mathbf{80.8}_{\color{red}+1.6}$ \\
           & DACAT      & $94.1\pm4.3$          & $\mathbf{89.2}$ & $88.5$          & $80.6$          \\
           & Ours       & $\mathbf{94.5\pm3.6}_{\color{red}+0.4}$ & $88.4_{\color{green}-0.8}$          & $\mathbf{90.2}_{\color{red}+1.7}$ & $\mathbf{81.4}_{\color{red}+0.8}$ \\
AutoLaparo & BNPitfalls & $86.8\pm1.5$          & $78.2$          & $72.0$          & $64.2$          \\
           & Ours       & $\mathbf{86.9\pm9.2}_{\color{red}+0.1}$ & $\mathbf{80.5}_{\color{red}+2.3}$ & $\mathbf{73.0}_{\color{red}+1.0}$ & $\mathbf{64.4}_{\color{red}+0.2}$ \\
           & DACAT      & $\mathbf{87.8\pm7.6}$ & $78.5$          & $75.0$          & $66.9$          \\
           & Ours       & ${87.6\pm7.8}_{\color{green}-0.2}$         & $\mathbf{84.5}_{\color{red}+6.0}$ & $\mathbf{76.4}_{\color{red}+1.4}$ & $\mathbf{67.9}_{\color{red}+1.0}$ 
\end{tblr}
}
\caption{The improvement of SOTAs for recognition task using $\mathcal{D}$.}
\label{tab:tab5_abla_5_2}
\end{table}

\begin{figure}[t]
\centerline{\includegraphics[width=\columnwidth]{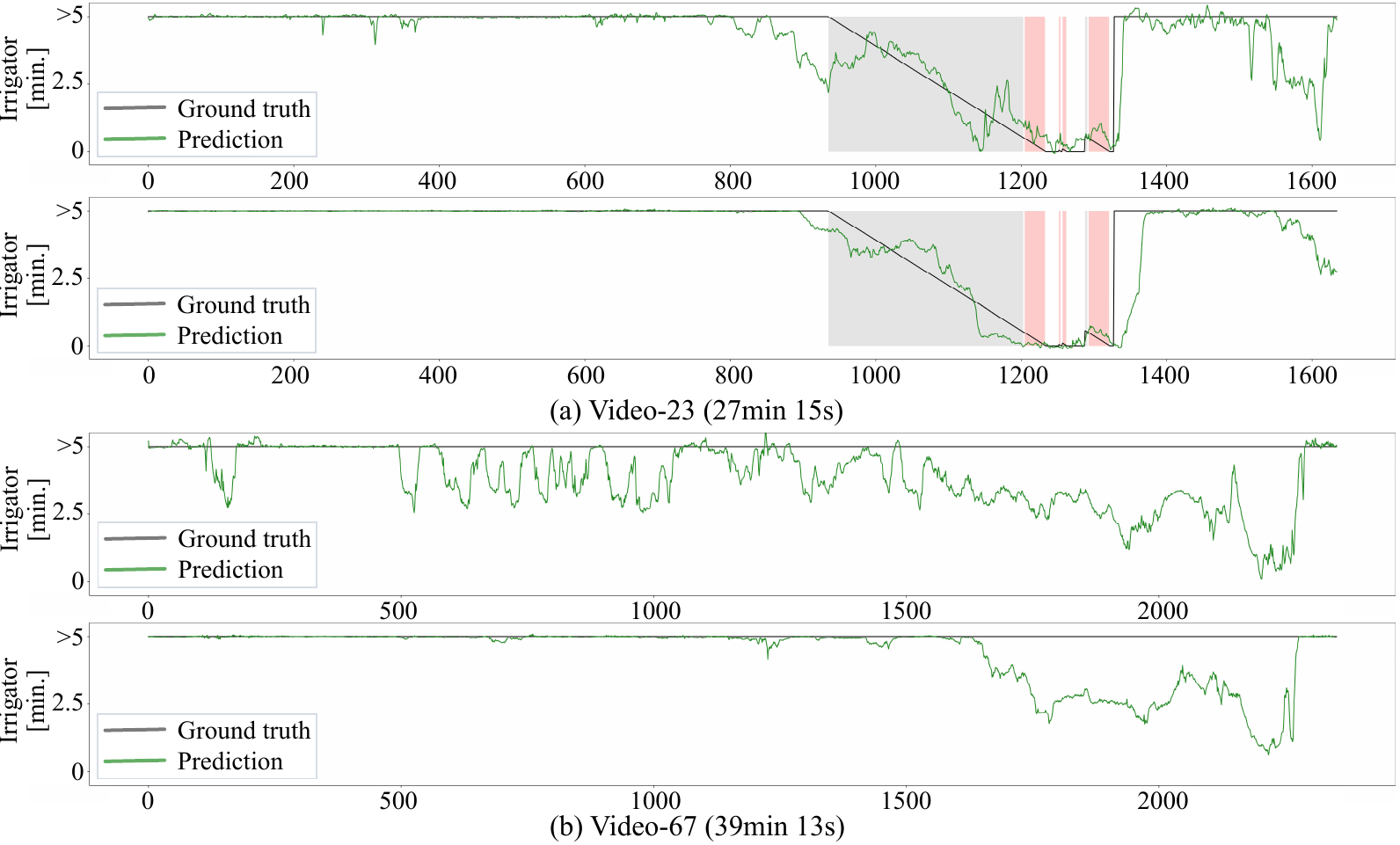}}
\caption{Prediction of usage trend for rarely unused instrument. (a) shows a case where the irrigator is used during surgery. (b) is a case where the irrigator is not used, yet both methods predict the usage trend. In each sub-figure, the top row represents the SOTA method BNPitfalls, and the bottom row represents ours.} 
\label{fig:fig9_con}
\end{figure}

%% file: sec/4_discussion.tex
\section{Discussion}
\textbf{Incorrect Usage Trend Anticipation.}
We observe some phenomena during the analysis of the experimental results. For the usage of the irrigator, the majority of patients undergoing cholecystectomy require it to clean the surgical area, remove bile, blood, and other secretions. However, a small number of patients do not require it during the procedure, possibly due to optimal surgical visualization, which enables the surgeon to proceed without an irrigator. 
As shown in Figure~\ref{fig:fig9_con}, (a) demonstrates a case where the irrigator is used, and (b) shows a case where it is not used. In (a), our method accurately anticipates the usage of the instrument, while in (b), both methods incorrectly predict the trend of usage. Regarding whether the instrument is needed or not, due to the limited data, the model cannot learn the criteria for its usage, as it depends on the surgeon and the specific circumstances of the patient. In the future, combining surgical fundamental models could improve the accuracy of instrument usage anticipation.

\section{Conclusion}
Building on methods in the surgical workflow analysis domain, we found that deterministic methods struggle to handle individual patient differences due to issues like collapse into dominant patterns. We propose CoStoDet-DDPM, the first method to utilize DDPM as a feature enhancer. It provides uncertainty-aware feature enhancement to deterministic models, while the task branch embeds clinical knowledge and enforces consistency constraints on the DDPM branch.
Ultimately, during the inference stage, the DDPM branch can be discarded to meet real-time requirement, making our approach the first to leverage DDPM model capabilities in this field. 
Our method demonstrates superior performance in both anticipation and regression tasks, as well as across different datasets. Through various ablation experiments, we have clearly and rigorously explored the roles of stochastic and deterministic models. In the future, we will integrate surgical foundational models to further enhance algorithm performance.

%% file: sec/X_suppl.tex
\clearpage
\setcounter{page}{1}
\maketitlesupplementary

\section{More Ablation Study}

\subsection{Observation Encoder of DDPM}
We explored the impact of different observation encoders in the DDPM. As shown in Table~\ref{tab_supple_1}, it is evident that incorporating temporal processing with LSTM, using longer sequences, and simultaneously performing both tool and phase tasks are beneficial for the results. Most importantly, incorporating the task loss term $\mathcal{L}_{Task}$ significantly improves performance.

\begin{table}[ht]
\centering
\resizebox{\linewidth}{!}{
\begin{tblr}{
  cells = {c},
  vline{3} = {-}{},
  hline{1-2,8} = {-}{},
  hline{1,8} = 1pt
}
Encoder  & seq & Tool           & Phase          & $Smooth$      \\
ResNet18-GN   & $32$  & $1.35/0.63/2.61$ & -              & -           \\
ConvNeXt      & $32$  & $1.17/0.49/2.46$ & -              & -           \\
ConvNeXt      & $64$  & $1.02/0.46/2.16$ & -              & -           \\
ConvNeXt-LSTM & $64$  & $0.97/0.29/2.05$ & -     & -  \\
ConvNeXt-LSTM & $64$  & $0.97/0.34/2.03$ & $0.63/0.19/1.05$ & $0.133+0.079$ \\
Ours ($w/$ $\mathcal{L}_{Task}$)         & $64$  & $0.93/0.34/1.75$ & $0.62/0.23/0.86$ & $0.095+0.077$ 
\end{tblr}
}
\caption{The impact of different observation encoders on DDPM performance. The results are evaluated on Cholec80 with \( h = 5 \) min. The columns present \( wMAE/outMAE/eMAE \), while \( Smooth \) is reported as \( Smooth_{Tool} + Smooth_{Phase} \). All inferences use DDIM with 16 steps. ResNet18-GN denotes a model where Batch Normalization is replaced with Group Normalization. }
\label{tab_supple_1}
\end{table}

\begin{table}[ht]
\centering
\resizebox{\linewidth}{!}{
\begin{tblr}{
  cells = {c},
  vline{3} = {-}{},
  hline{1-2,8} = {-}{},
  hline{4,6} = {1-5}{dashed},
  hline{1,8} = 1pt
}
Output Branch & $\lambda$ & Tool               & Phase              & $Smooth$               \\
$\mathcal{D}$   & $1$                 & $0.95/0.31/2.04$          & $0.63/0.20/1.03$          & $0.093+0.073$          \\
$\mathcal{T}$       & $1$                 & $0.94/0.32/2.04$          & $0.61/0.20/1.03$          & $0.031+0.026$          \\
$\mathcal{D}$   & $16$                & $\mathbf{0.91}/0.36/1.83$          & $\mathbf{0.58}/0.21/$$\mathbf{0.83}$ & $0.057+0.044$          \\
$\mathcal{T}$       & $16$                & $\mathbf{0.91}/0.35/1.85$          & $\mathbf{0.58}/0.20/0.85$ &$ 0.028+0.021$          \\
$\mathcal{D}$   & $32$                & $0.93/0.34/\mathbf{1.75}$ & $0.62/0.23/0.86$          & $0.095+0.077$          \\
$\mathcal{T}$       & $32$                & $\mathbf{0.91}/\mathbf{0.29}/1.77$ & $0.59/\mathbf{0.18}/0.88$ & $\mathbf{0.025}+\mathbf{0.020}$ 
\end{tblr}
}
\caption{Comparison of results between \(\mathcal{D}\) and \(\mathcal{T}\) with different \(\lambda\). The evaluation is conducted on Cholec80 with \( h = 5 \) min. The columns report \( wMAE/outMAE/eMAE \), while $Smooth$ is presented as \( Smooth_{Tool} + Smooth_{Phase} \). All experiments use CoStoDet-DDPM, with \(\mathcal{D}\) inferences performed using DDIM with 16 steps, and the clip sequence length set to 64.}
\label{tab_supple_2}
\end{table}

\subsection{Historical Time Span of Labels in DDPM}

We also investigated the historical span \(\lambda\) of labels, as shown in the Table~\ref{tab_supple_2}. Due to memory and time constraints, we explored \(\lambda = 1\), \(16\), and \(32\). When \(\lambda = 1\), only the current frame label is used. The results show that incorporating historical information improves the anticipation performance, and the effects of \(\lambda = 16\) and \(\lambda = 32\) are nearly identical.

\subsection{Denoising Network Architecture and Conditional Feature Fusion}
\label{exp:condition-and-net}
\begin{figure}[t]
\centerline{\includegraphics[width=\columnwidth]{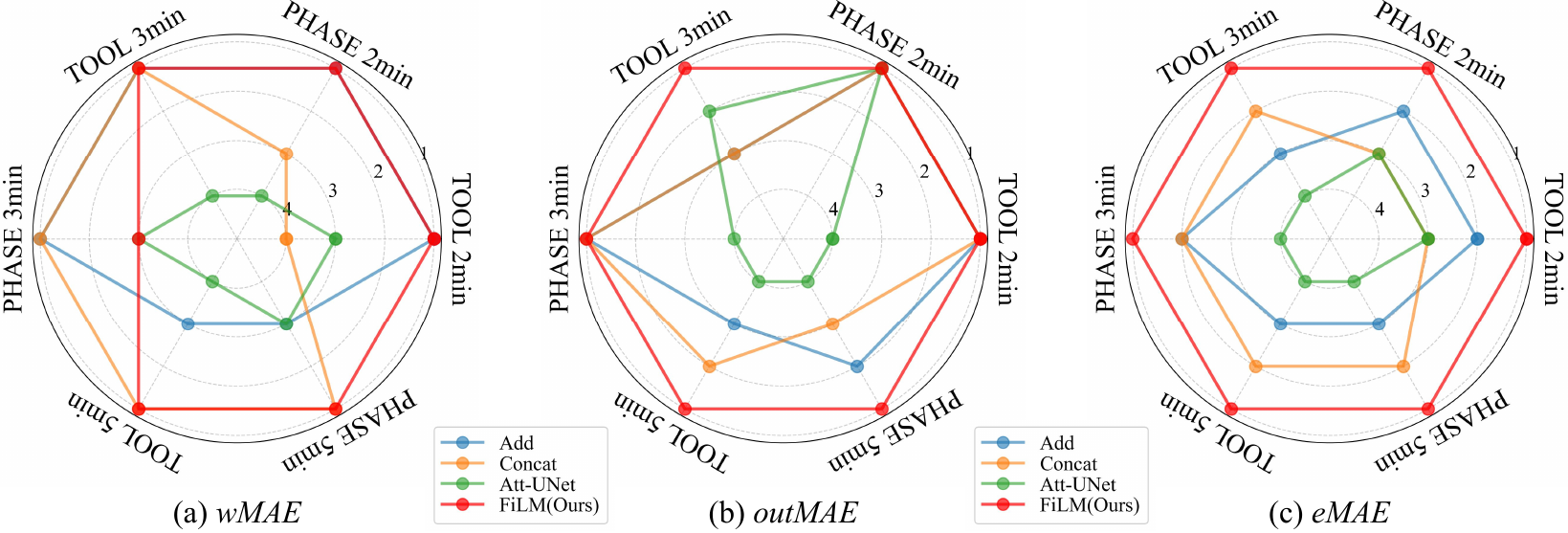}}
\caption{The impact of denoising network architecture and conditional feature fusion. 
The scale represents rankings, with detailed data provided in the Supplementary Materials.} 
\label{fig:fig6_aba2}
\end{figure}

We explored the impact of feature fusion methods and denoising network architectures within $\mathcal{D}$. Specifically, we investigated three approaches: (1) directly adding the conditional features at each layer (Add), (2) concatenating the conditional features with the noisy features before inputting them into the U-Net (Concat), and (3) using Attention U-Net (Att-UNet) as the denoising network. As shown in Figure~\ref{fig:fig6_aba2}, FiLM and U-Net are more effective for co-training, leading to improved performance.

\subsection{Concrete Experiment Results}
\label{sec:rationale}

We present more detailed results of two ablation experiments: Denoising Network Architecture and Conditional Feature Fusion~\cref{exp:condition-and-net}, as shown in Table~\ref{tab:SM1}, and Different Collaborative Learning Strategies~\cref{exp:co-training}, as shown in Table~\ref{tab:SM2}.

\begin{table*}
\centering
\begin{tblr}{
  cells = {c},
  cell{1}{1} = {r=3}{},
  cell{1}{2} = {c=6}{},
  cell{2}{2} = {c=3}{},
  cell{2}{5} = {c=3}{},
  vline{2} = {1-7}{},
  vline{5} = {2-7}{},
  hline{1,4,7-8} = {-}{},
  hline{2-3} = {2-7}{},
  hline{1,8} = 1pt
}
Method     & $wMAE$~$\downarrow$  / $outMAE$~$\downarrow$ / $eMAE$~$\downarrow$ &                &                &                &                &                \\
           & Tool                       &                &                & Phase          &                &                \\
           & $h=2$ min & $h=3$ min & $h=5$ min & $h=2$ min & $h=3$ min & $h=5$ min \\
Add        & $0.39/0.05/0.94$             & $0.56/0.13/1.31$ & $0.95/0.33/2.07$ & $0.27/0.05/0.58$ & $0.37/0.09/0.68$ & $0.60/0.19/0.99$ \\
Concat     & $0.41/0.05/0.99$             & $0.56/0.13/1.29$ & $0.91/0.30/1.94$ & $0.28/0.05/0.59$ & $0.37/0.09/0.68$ & $0.59/0.21/0.89$ \\
Att-UNet   & $0.40/0.06/0.99$             & $0.57/0.11/1.36$ & $0.96/0.36/2.08$ & $0.29/0.05/0.59$ & $0.39/0.10/0.72$ & $0.60/0.22/1.04$ \\
FiLM(Ours) & $0.39/0.05/0.93$             & $0.56/0.09/1.28$ & $0.91/0.29/1.77$ & $0.27/0.05/0.52$ & $0.39/0.09/0.67$ & $0.59/0.18/0.88$ 
\end{tblr}
\caption{The impact of denoising network architecture and conditional feature fusion.}
\label{tab:SM1}
\end{table*}

\begin{table*}
\centering
\begin{tblr}{
  cells = {c},
  cell{1}{1} = {r=3}{},
  cell{1}{2} = {c=6}{},
  cell{2}{2} = {c=3}{},
  cell{2}{5} = {c=3}{},
  vline{2} = {1-9}{},
  vline{5} = {2-9}{},
  hline{1,4,9-10} = {-}{},
  hline{2-3} = {2-7}{},
  hline{1,10} = 1pt
}
Method     & $wMAE$~$\downarrow$  / $outMAE$~$\downarrow$ / $eMAE$~$\downarrow$ &                &                &                &                &                \\
           & Tool                       &                &                & Phase          &                &                \\
           & $h=2$ min & $h=3$ min & $h=5$ min & $h=2$ min & $h=3$ min & $h=5$ min \\
BNP        & $0.40/0.06/1.06$             & $0.58/0.13/1.42$ & $0.96/0.41/2.11$ & $0.29/0.06/0.63$ & $0.39/0.10/0.78$ & $0.61/0.24/0.99$ \\
MaskAE     & $0.41/0.13/0.97$             & $0.64/0.26/1.47$ & $1.17/0.66/2.27$ & $0.29/0.10/0.62$ & $0.45/0.20/0.76$ & 0.79/0.44/1.21 \\
CURL       & $0.41/0.07/1.01$             & $0.58/0.16/1.44$ & $1.01/0.52/2.05 $& $0.27/0.06/0.55$ & $0.39/0.11/0.74$ & $0.63/0.32/0.91$ \\
GC         & $0.40/0.08/1.03$             & $0.59/0.19/1.35$ & $0.98/0.44/2.04$ & $0.29/0.08/0.60$ & $0.39/0.13/0.66$ & $0.60/0.24/0.93$ \\
RandomMask & $0.41/0.05/0.99$             & $0.56/0.11/1.29$ & $0.91/0.28/1.86$ & $0.30/0.06/0.60$ & $0.38/0.09/0.70$ & $0.60/0.17/0.90$ \\
FiLM(Ours) & $0.39/0.05/0.93$             & $0.56/0.09/1.28$ & $0.91/0.29/1.77$ & $0.27/0.05/0.52$ & $0.39/0.09/0.67$ & $0.59/0.18/0.88$ 
\end{tblr}
\caption{The impact of co-training methods.}
\label{tab:SM2}
\end{table*}

\section{More Discussion}
\subsection{Sensitivity to Hyperparameter for Recognition}
For phase recognition, we find that CoStoDet-DDPM is sensitive to the learning rate ($lr$) and weight decay ($wd$) during training, as shown in Table~\ref{tab:tab6_con}. For BNPitfalls, a higher learning rate than the original one is required to fully leverage the effect of \( \mathcal{D} \). For DACAT, the original $lr$ and $wd$ settings are sufficient. Due to time and resource constraints, we have not yet fully explored the optimal parameters for the recognition task, and further investigation is needed in the future.

\begin{table}[t]
\centering
\resizebox{\linewidth}{!}{
\begin{tblr}{
  cells = {c},
  vline{2-3} = {-}{},
  hline{1-2,6,9} = {-}{},
  hline{3,7} = {-}{dashed},
  hline{1,9} = 1pt
}
Methods     & $lr$, $wd$     & Accuracy$~\uparrow$ & Precision~$\uparrow$ & Recall$~\uparrow$ & Jaccard$~\uparrow$   \\
BNPitfalls & $1e-4$,$1e-2$ & $93.7\pm4.1$          & $88.7$          & $87.7$          & $79.2$           \\
$w/~\mathcal{D}$      & $1e-4$,~$1e-2$ & $92.2\pm6.6$          & $87.0$          & $86.6$          & $77.2$           \\
$w/~\mathcal{D}$      & $1e-3$,~$1e-2$ & $\mathbf{94.1\pm3.5}$ & $\mathbf{88.9}$ & $\mathbf{89.2}$ & $\mathbf{80.8}$  \\
$w/~\mathcal{D}$      & $1e-3$,~$1e-6$ & $91.3\pm6.7$          & $84.0$         & $86.3$          & $74.7$           \\
DACAT      & $1e-5$,~$1e-2$ & $94.1\pm4.3$          & $\mathbf{89.2}$          & $88.5$          & $80.6$           \\
$w/~\mathcal{D}$      & $1e-5$,~$1e-2$ & $\mathbf{94.5\pm3.6}$ & $88.4$ & $\mathbf{90.2}$ & $\mathbf{81.4~}$ \\
$w/~\mathcal{D}$      & $1e-5$,~$1e-6$ &$ 94.4\pm3.6$          & $88.1$          & $89.8$          & $80.9$           
\end{tblr}
}
\caption{The impact of different learning rates ($lr$) and weight decay ($wd$) for recognition results on Cholec80.}
\label{tab:tab6_con}
\end{table}